\documentclass[conference, 9pt]{IEEEtran}
\IEEEoverridecommandlockouts
\usepackage{cite}
\usepackage{amsmath,amssymb,amsfonts}
\usepackage{graphicx}
\usepackage{textcomp}
\usepackage{xcolor}
\usepackage{ dsfont }
\usepackage{booktabs}
\usepackage{tablefootnote}
\usepackage{pbox}
\usepackage{enumitem}
\usepackage{pifont}
\usepackage{tikz}
\usepackage{inconsolata}
\usepackage{algorithm}
\usepackage{algorithmicx}
\usepackage{lipsum}
\usepackage[noend]{algpseudocode}

\algnewcommand\algorithmicforeach{\textbf{for each}}
\algdef{S}[FOR]{ForEach}[1]{\algorithmicforeach\ #1\ \algorithmicdo}

\algnewcommand{\IIf}[1]{\State\algorithmicif\ #1\  \algorithmicthen}
\algnewcommand{\EndIIf}{\unskip}
\usepackage{mathrsfs}

\usepackage{multirow}
\usepackage{longtable}

\usepackage{hyperref}

\def\BibTeX{{\rm B\kern-.05em{\sc i\kern-.025em b}\kern-.08em
    T\kern-.1667em\lower.7ex\hbox{E}\kern-.125emX}}

\renewcommand{\IEEEbibitemsep}{0pt plus 0.5pt}
\makeatletter
\IEEEtriggercmd{\reset@font\normalfont\fontsize{7.8pt}{8pt}\selectfont}
\makeatother
\IEEEtriggeratref{1}
\newcommand\encircle[1]{%
  \tikz[baseline=(X.base)] 
    \node (X) [draw, shape=circle, inner sep=-1, fill=black, text=white] {\strut #1};%
}

\usepackage{xspace}
\newcommand{\Design}{\textit{$\mathsf{DistHD}$\xspace}}    

\begin{document}

\title{DistHD: A Learner-Aware Dynamic Encoding Method for Hyperdimensional Classification\vspace{-3mm}}

\author{
    \IEEEauthorblockN{Junyao Wang$^\dag$, Sitao Huang$^\S$, Mohsen Imani$^\dag$}
    \IEEEauthorblockA{\textit{$^\dag$ Department of Computer Science, University of California, Irvine, CA, United States}\\
    \textit{$^\S$ Department of Electrical Engineering and Computer Science, University of California, Irvine, CA, United States}
    \\\textit{\{junyaow4, sitaoh, m.imani\}}@uci.edu\vspace{-3mm}}%\vspace{-6mm}
}
\maketitle

\begin{abstract}
The Internet of Things (IoT) has become an emerging trend that connects heterogeneous devices and enables them with new capabilities. Many applications exploit machine learning methodology to dissect collected data, and \textit{edge computing} was introduced to enhance the efficiency and scalability in resource-constrained computing environments. Unfortunately, popular deep learning algorithms involve intensive computations that are overcomplicated for edge devices. Brain-inspired Hyperdimensional Computing (HDC) has been considered a promising approach to address this issue. 
However, existing HDC methods use static encoders, and thus require extremely high dimensionality and hundreds of training iterations to achieve reasonable accuracy. This results in a huge loss of efficiency and severely impedes the application of HDC algorithms in power-limited machines. In this paper, we propose $\Design$, a novel HDC framework with a unique dynamic encoding technique consisting of two parts: \textit{top-2 classification} and \textit{dimension regeneration}. Our \textit{top-2 classification} provides top-2 labels for each data sample based on cosine similarity, 
and \textit{dimension regeneration} identifies and regenerates dimensions that mislead the classification and reduce the learning quality. The highly parallel algorithm of $\Design$ effectively accelerates the learning process and achieves the desired accuracy with considerably lower dimensionality. Our evaluation on a wide range of practical classification tasks shows that $\Design$ is capable of achieving on average $2.12\%$ higher accuracy than state-of-the-art (SOTA) HDC approaches while reducing dimensionality by $8.0 \times$. It delivers $5.97\times$ faster training and $8.09\times$ faster inference than SOTA learning algorithms. Additionally, the holographic distribution of patterns in high dimensional space provides $\Design$ with $12.90\times$ higher robustness against hardware errors than SOTA DNNs. $\Design$ has been open-sourced to enable future research in this field. \footnote{$\Design$ source code: \url{https://github.com/jwang235/DistHD}}
\end{abstract}


\section{Introduction}\label{sec:intro}
The Internet of Things (IoT) has recently become an emerging trend for its extraordinary potential to connect various heterogeneous smart sensors and devices and enable them with new capabilities. Many IoT applications exploit machine learning (ML) algorithms to dissect collected data and perform learning and cognitive tasks. However, the excellent learning quality of popular ML approaches, including deep neural networks (DNNs), often comes at the expense of high computational and memory requirements, involving millions of parameters that need to be iteratively refined over multiple time periods~\cite{chen2019deep}. One common approach is to leverage cloud computing by sending data from the network edge to the centralized location in the cloud. %~\cite{zhu2020toward}. 
Unfortunately, this potential solution results in significant efficiency loss, multiple scalability issues, and serious privacy concerns~\cite{shi2016edge}. \textit{Edge computing}, a novel computing paradigm performing calculations in proximity to data sources, has since been introduced to address these issues. However, accommodating the high resource requisite of traditional learning methodologies on less powerful computing platforms remains a critical challenge to surmount~\cite{chen2019deep}. %, najafabadi2015deep}. 
Considering the increasingly massive amount of information nowadays, the power and memory limitations of embedded devices, and the potential instabilities of IoT systems, a more lightweight, efficient, and robust learning algorithm is of absolute necessity~\cite{pan2017future}. 
\begin{figure}[!t]
\centering
%\vspace{-5mm}
\includegraphics[width=\linewidth]{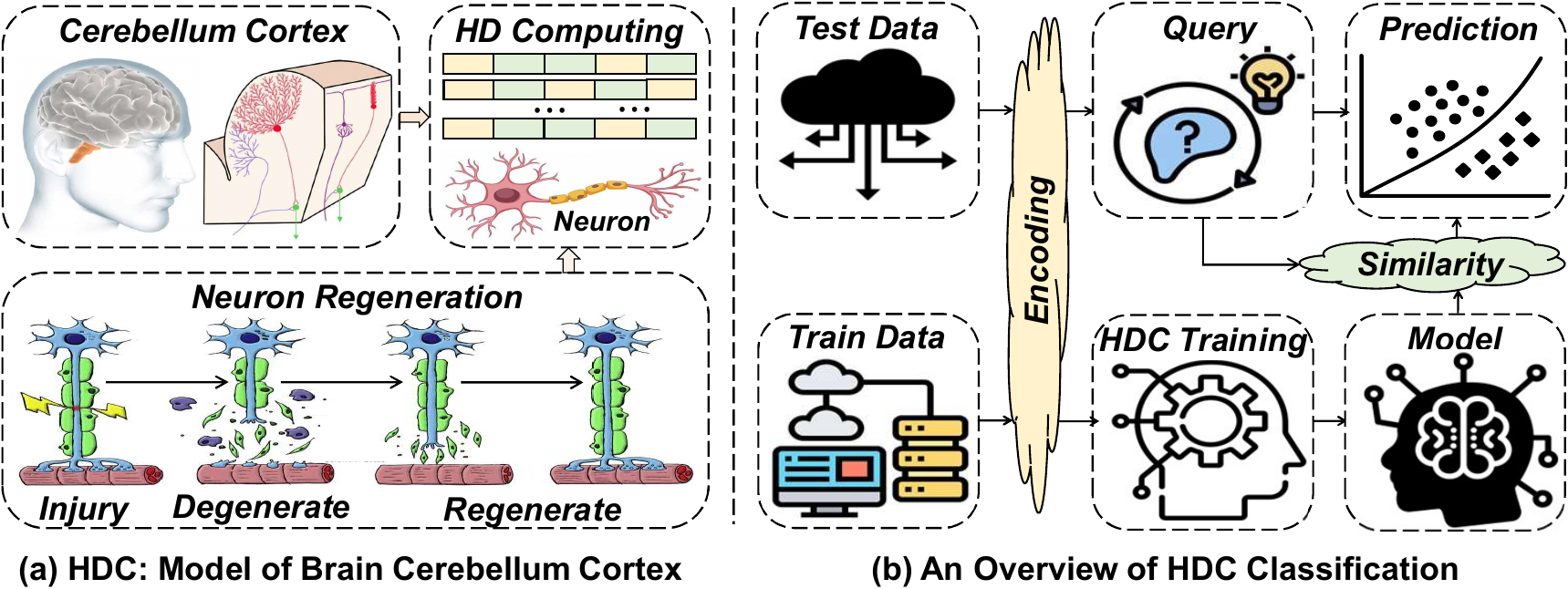}
\vspace{-5.5mm}
\caption{An Overview of Brain Cerebellum Cortex and HDC Classification}
\vspace{-2mm}
\label{fig: overview}  
\end{figure}

% \begin{figure}[!t]
% \centering
% %\vspace{-3mm}
% \includegraphics[width=\linewidth]{figures/motivation-crop.pdf}
% \vspace{-6mm}
% \caption{Motivation for Dynamic Encoding and Top-2 Classification}
% \vspace{-6mm}
% \label{fig: motivation}  
% \end{figure}

\begin{figure}[!t]
\centering
%\vspace{-3mm}
\includegraphics[width=\linewidth]{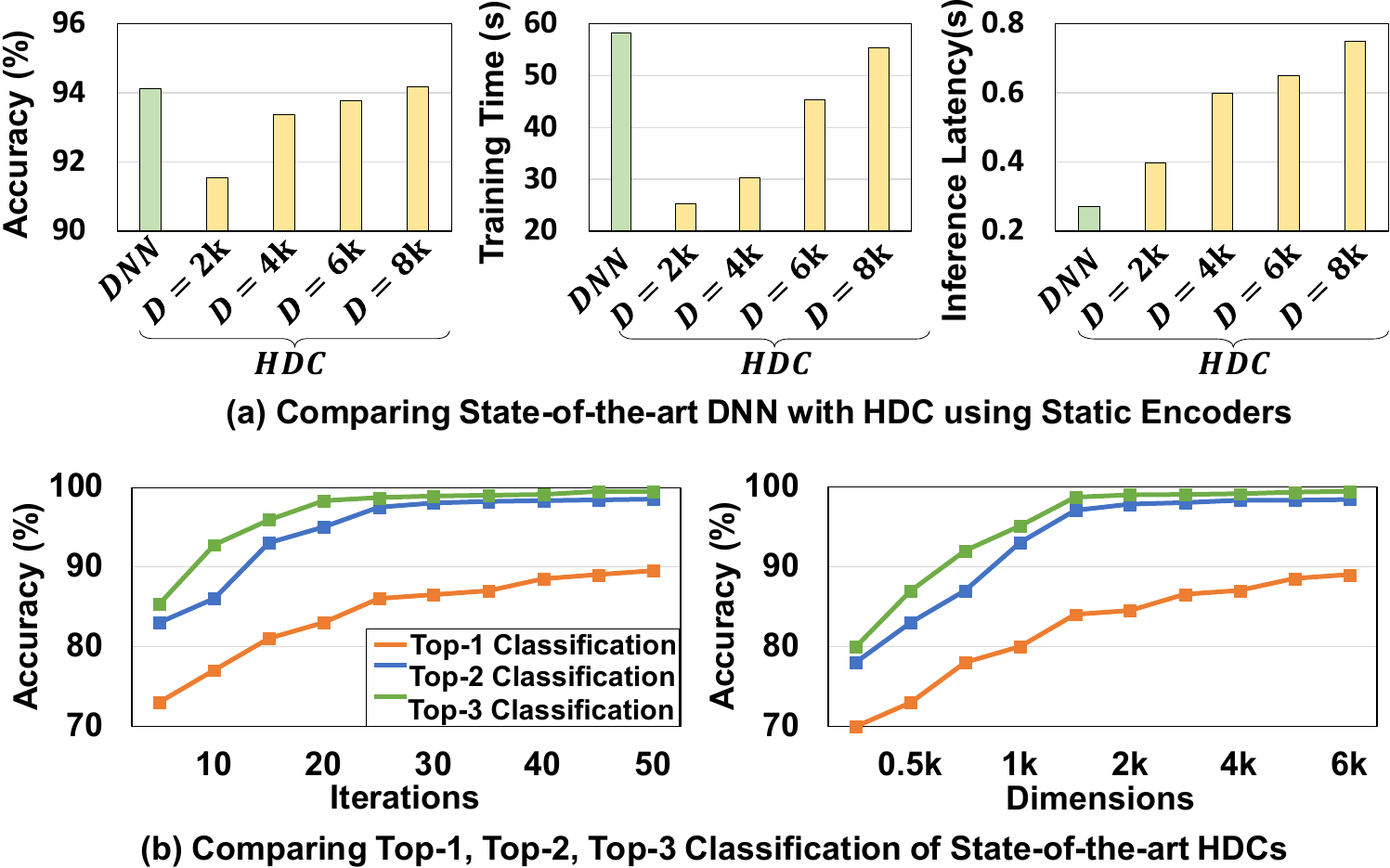}
\vspace{-5.5mm}
\caption{Motivation for Dynamic Encoding and Top-2 Classification}
\vspace{-5mm}
\label{fig: motivation}  
\end{figure}

In contrast to traditional artificial intelligence methodologies, HDC is considered a promising learning approach for less powerful computing platforms for its (\romannumeral 1) high computational efficiency ensuring real-time learning~\cite{ge2020classification}, (\romannumeral 2) strong robustness to noise -- a key strength for IoT systems~\cite{imani2019framework}, and (\romannumeral 3) lightweight hardware implementation enabling efficient execution on edge device~\cite{rahimi2016robust}. As demonstrated in Fig. \ref{fig: overview}(a), HDC is motivated by the neuroscience observation that the cerebellum cortex in the human brain is capable of effortlessly and efficiently processing memory, perception, and cognition information without much concern for noisy or broken neuron cells. %~\cite{yilmaz2015symbolic}. %, kanerva2009hyperdimensional}. 
Closely mimicking the information representation and memorization functionalities of human brains, HDC encodes low-dimensional inputs to \textit{hypervectors} with $10^4$ or more elements to perform various learning tasks~\cite{zou2021scalable} as shown in Fig. \ref{fig: overview}(b). HDC then conducts highly parallel and well-defined
operations and has been proven to achieve high-quality results in classification and regression learning tasks with comparable accuracy to state-of-the-art (SOTA) DNNs and SVMs. Additionally, the notably faster convergence and higher efficiency offered by HDC provide a powerful solution for today's embedded devices with limited storage, battery, and resources~\cite{ge2020classification, rahimi2016robust, imani2019framework}. 

Despite the enormous success in the development of HDC, as demonstrated in Fig. \ref{fig: motivation}(a), existing HDC algorithms require extremely high dimensionality ($D$) to outperform DNNs. Consequently, not only is the learning efficiency considerably lowered with large numbers of unnecessary computations involved, but the system efficiency is also compromised due to the increased data size and communication cost~\cite{imani2019framework}. This severely impedes the feasibility and scalability of HDC in resource-constrained computing devices, especially for learning tasks including massive amounts of data and requiring real-time analysis. We observed that one of the main causes is that the encoding module of existing HDC approaches lacks the capability to utilize and adapt to information learned during the training process. On contrary, as demonstrated in Fig.\ref{fig: overview}(a), neurons in human brains dynamically change and regenerate all the time and provide more useful functionality when they learn new information~\cite{andersen2003aging}. While the goal of HDC is to exploit the high-dimensionality of randomly generated base hypervectors to represent the information as a pattern of neural activity, it remains challenging for existing HDC algorithms to support a similar behavior as brain neural regeneration.

One interesting observation of SOTA HDC approaches is that they provide considerably higher accuracy and faster convergence for $\textit{top-2 classification}$ than top-1 classification, as shown in Fig. \ref{fig: motivation}(b). Here we define a \textit{top-k classification} for a given data point as \textit{correct} if the true label is one of the \textit{k} most similar classes selected. Additionally, the accuracy difference between top-2-classification and top-3 classification is noticeably smaller than that between top-1 classification and top-2-classification. Based on this observation, in this paper, we propose $\Design$, a new HDC framework with an innovative encoding technique that utilizes and adapts to information learned from every training iteration. $\Design$ aims at identifying dimensions that mislead the classification and decrease the learning accuracy, and regenerating them for a more positive impact on the learning quality. The main contributions of the paper are listed below:

\begin{itemize}[leftmargin = *]
\item We propose a novel dynamic encoding technique for HDC combining \textit{top-2 classification} and \textit{dimension regeneration} optimizations. To the best of our knowledge, $\Design$ is the first HDC algorithm with a dynamic encoding module that identifies and regenerates dimensions hurting the classification accuracy to enhance learning quality. $\Design$ achieves on average $2.12\%$ higher classification accuracy than SOTA HDCs while reducing the number of needed dimensions by $8.0\times$. This ensures accurate performance for classification tasks on resource-constrained devices. 
\item The highly parallel and matrix-wise operations of $\Design$ ensures a considerable acceleration for performing classification tasks in both high-performance and resource-constrained computing environments. $\Design$ delivers $5.97\times$ faster training than SOTA DNNs and $8.09\times$ faster inference than SOTA HDC algorithms. 
%demonstrates a $5.97\times$ speedup in training compared to SOTA DNNs and an $8.09\times$ speedup in inference compared to SOTA HDC algorithms. %This ensures a considerable acceleration for performing classification tasks in both high-performance and resource-constrained computing environments.
\item The holographic distribution of patterns in high-dimension space enables $\Design$ with notably higher robustness. Our proposed model demonstrates on average $12.90\times$ higher robustness against hardware errors than SOTA DNNs. This ensures the effective execution of classification tasks on noisy IoT devices. 
% \item $\Design$ delivers on average $3.77\%$ higher classification accuracy than state-of-the-art HDC approaches, and demonstrates a $\times$ and $\times$ faster performance in training and inference, respectively. This ensures efficient and accurate performance for classification tasks on resource-constrained devices. 
% \item $\Design$ achieved comparable accuracy with state-of-the-art DNNs and $12.9\times$ robustness against hardware noise than state-of-the-art DNNs, and achieves comparably high accuracy. This ensures real-time execution of classification tasks on edge devices. 
% \item $\Design$ demonstrates higher accuracy than state-of-the-art HDC algorithms using the same dimensionality. This ensures an accurate and effective execution of classification tasks on resource-constrained devices. 
% \item Our evaluation on FPGA shows that $\Design$ is hardware friendly and can work under various bitwidths; it achieves higher energy efficiency than on CPU baseline. 
\end{itemize}

The rest of the paper is organized as follows: in Section \ref{sec:background}, we briefly introduce recent works in edge-based learning and HDC. We then explain our proposed methodology in detail in Section \ref{sec:method}, and evaluate our model in terms of accuracy, efficiency, and robustness against noise in Section \ref {sec:result}. We conclude our work in Section \ref{sec:conclusion}.

% \begin{figure}[!t]
% \centering
% %\vspace{-5mm}
% \includegraphics[width=\linewidth]{figures/overview.pdf}
% %\includegraphics[width=\textwidth]{figures/partial.pdf}
% \vspace{-7mm}
% \caption{An Overview of Brain Cerebellum Cortex and HDC Classification}
% %\vspace{-6mm}
% \label{fig: overview}  
% \end{figure}

% \begin{figure}[!t]
% \centering
% \vspace{-3mm}
% \includegraphics[width=\linewidth]{figures/motivation.pdf}
% \vspace{-7mm}
% \caption{Motivation For Utilizing Top2 Classifier}
% \vspace{-6mm}
% \label{fig: motivation}  
% \end{figure}

\section{Related Works} \label{sec:background}
\subsection{IoT and Edge-based Learning}
The rapid development of IoT has engendered %traditional cloud computing is no longer sufficient for the diverse needs of today's intelligent society~\cite{cao2020overview}. 
a number of innovative works %have been engendered 
on the feasibility and scalability of edge-based learning. 
Prior works have demonstrated that machine learning algorithms (including DNNs) can possibly be customized for learning on edge computing devices. Various frameworks and libraries have been developed for learning on the edge, including TinyML~\cite{warden2019tinyml}, TensorFlow Lite~\cite{david2021tensorflow}, edge-ml~\cite{sakr2020machine}, X-Cube-AI~\cite{xcubeai}, etc. These frameworks are all machine learning or deep learning based tools. Many of these learning methods require a large number of training samples and long training cycles, and may not meet the tight constraints of ultra-low-power edge platforms. 
On the other hand, people propose techniques that improve learning efficiency on the edge, leveraging learning structures and target platform properties. Representative examples include split computing~\cite{ko2018edge}, federated learning~\cite{bonawitz2017practical, li2020federated}, knowledge distillation~\cite{luo2022keepedge}, etc. These techniques are orthogonal to our method and can potentially be integrated with our learning solution.

\subsection{Hyperdimensional Computing}
 %HDC has been consistently considered as a promising learning approach in resource-constrained computing environments for its high computational efficiency, strong robustness against noise, and lightweight hardware implementation. 
 Prior studies have exhibited enormous success in various applications of HDCs, such as brain-like reasoning~\cite{poduval2022graphd}, bio-signal processing~\cite{burrello2019laelaps}, and human-activity recognition~\cite{kim2018efficient}. A few endeavors have been made towards developing novel architecture to accelerate HDC inference tasks~\cite{imani2017exploring}. However, popular HDC methods use pre-generated static encoders %that are never updated during the training process. Consequently, 
 and thus require extremely high dimensionality to achieve acceptable accuracy~\cite{rahimi2007random}. %Consequently, it involves intensive computations and severely impeding the application of HDC approaches in resource-constrained devices. 
 NeuralHD~\cite{zou2021scalable}, a recently proposed dynamic encoding approach, successfully compressed dimensionality by eliminating dimensions with minor impacts on distinguishing patterns. However, its proposed model takes a significantly longer time than SOTA HDC algorithms~\cite{rahimi2016robust} to reach convergence, and it lacks an effective technique to enhance its encoding module with information learned during the training process. %to enhance learning quality. 
 In contrast, we propose $\Design$, aiming at fully exploiting information learned from each iteration and achieving adequate accuracy with much faster convergence and lower dimensionality.

\section{Methodology} \label{sec:method}

% \begin{figure*}[t]
% %\vspace{-5mm}
%   \centering
%   {\includegraphics[width=\textwidth]{figures/workflow-crop.pdf}}
%   % {\includegraphics[width=\textwidth]{figures/flow.pdf}}
%   %\vspace{-4mm}
%   \caption{An Overview of $\Design$ Workflow }
%  \vspace{-3mm}
%   \label{fig: flow} 
% \end{figure*}

\begin{figure*}[t]
%\vspace{-5mm}
  \centering
  {\includegraphics[width=\textwidth]{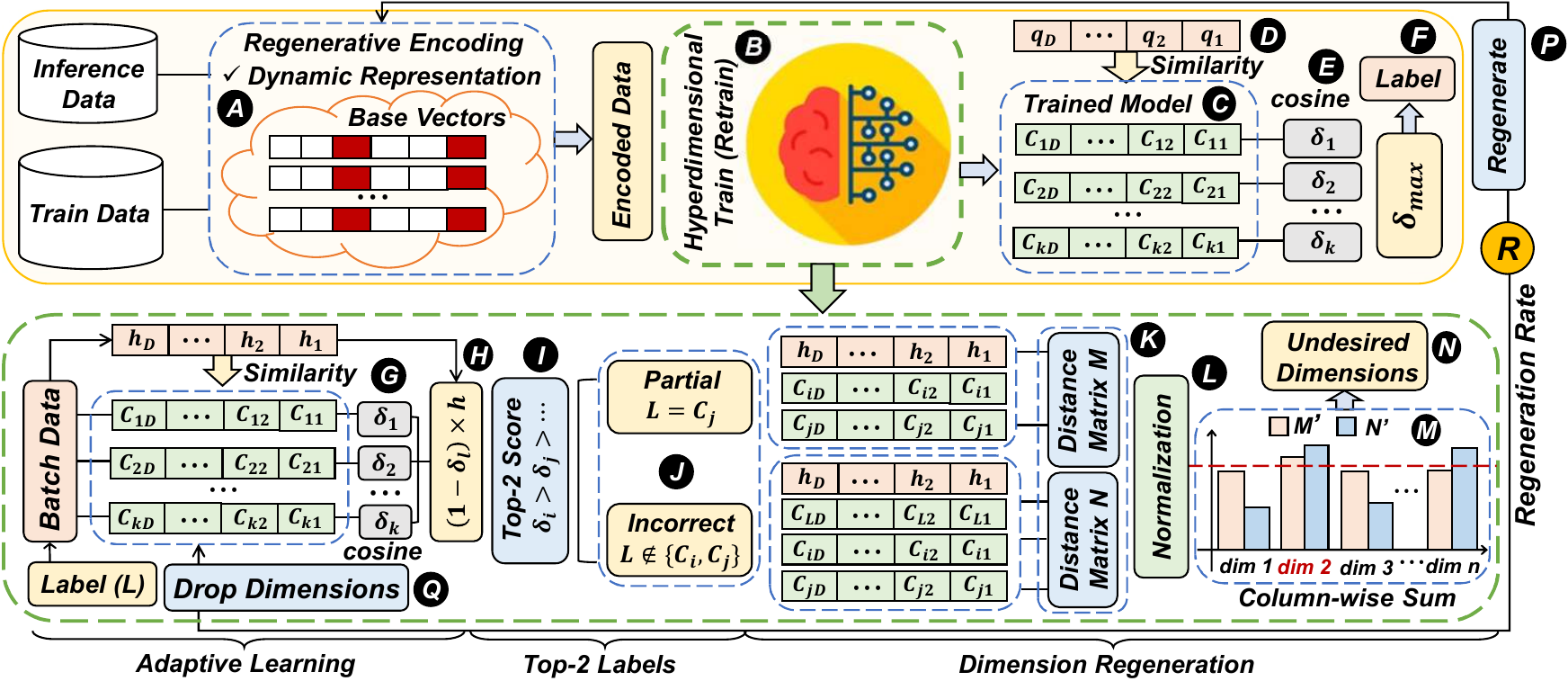}}
  % {\includegraphics[width=\textwidth]{figures/flow.pdf}}
  \vspace{-6mm}
  \caption{An Overview of $\Design$ Workflow }
 \vspace{-4mm}
  \label{fig: flow} 
\end{figure*}

Popular HDC algorithms use static encoding techniques where the pre-generated base vectors lack the capacity to adapt to information learned during the training process. The goal of our proposed $\Design$ is to effectively exploit information learned from each training iteration to identify dimensions that reduce the learning quality and regenerate them. % to enhance the classification accuracy. 
As demonstrated in Fig. \ref{fig: flow}, $\Design$ starts with encoding data points into high-dimensional space with existing encoding methods depending on the data type $(\encircle{A})$. $\Design$ then conducts two innovative steps, \textit{top-2 classification} and \textit{dimension regeneration}, to enable its encoding module and base vectors with adaptivity to each partially trained model. In each iteration of \textit{top-2 classification}, $\Design$ first applies a highly efficient adaptive learning algorithm over the encoded data $(\encircle{B})$, and then utilizes the partially trained model to compute the top two most similar classes for each data point $(\encircle{I})$. In \textit{dimension regeneration}, we calculate two distance matrices $(\encircle{K})$ based on results from the \textit{top-2 classification}, and identify $(\encircle{N})$ and eliminate $(\encircle{Q})$ undesired dimensions that mislead the classification tasks. To further improve the learning quality, we regenerate those dimensions $(\encircle{P})$ for a more positive impact on the classification. %$\Design$ eventually obtains a learning system where most of the dimensions positively impact learning quality. 
Note that all the operations in this section can be done in a highly parallel matrix-wise way, as multiple training samples can be grouped into a matrix of row hypervectors. 

\subsection{HDC Preliminaries}
Motivated by the high-dimensional information representation and memorization functionalities in human brains, %Motivated by the neuroscience observation that the human cerebellum cortex efficiently and effortlessly performs memory and cognition functions with operations on high-dimensional representations of data,
HDC maps inputs onto hyperdimensional space as \textit{hypervectors} $(\encircle{A})$, each of which contains thousands of elements. %Each dimension of hypervectors abstractly models a neuron functionality in performing cognition tasks~\cite{zou2021scalable}. Fig. \ref{fig: overview}(b) provides an overview of hyperdimensional classification. The first step of HDC algorithms is to \textit{encode} inputs into \textit{hypervectors} of high-dimensional space with an encoding technique depending on data types~\cite{rahimi2016robust}. 
One unique property of the hyperdimensional space is the existence of a large number of nearly orthogonal hypervectors, enabling highly parallel operations such as similarity calculations, bundlings, and bindings. Mathematically, consider random bipolar hypervectors ${\mathcal{H}}_1$ and ${\mathcal{H}}_2$ with dimension $\mathcal D$, i.e., ${\mathcal{H}}_1, {\mathcal{H}}_2 \in \{-1,1\}^{\mathcal D}$, when $\mathcal D$ is large enough, the dot product ${\mathcal{H}}_1\cdot  {\mathcal{H}}_2\approx 0$. %$\delta(\overrightarrow{\mathcal{H}}_1, \overrightarrow{\mathcal{H}}_2)\approx 0$ where $\delta$ denotes the cosine similarity. 
%  The orthogonality in HDC enables well-defined operations such as similarity calculations, bundling, and binding to be conducted in a highly parallel way. %The HDC encoding is based on the following operations:
\textbf{Similarity:} calculation of the distance between the query hypervector and the class hypervector (noted as $\delta(\cdot, \cdot)$). 
For real-valued hypervectors, %$\delta(\overrightarrow{\mathcal{H}_1}, \overrightarrow{\mathcal{H}_2})$
 a common measure is cosine similarity, i.e.
 \begin{equation}
    \label{eqn:cosine_similarity}
    \delta( {\mathcal H}, {\mathcal C_l})=\frac{ {\mathcal H} \cdot {\mathcal C_l}}{\| {\mathcal H}\|\cdot \| {\mathcal C_l}\|}=\frac{ {\mathcal H}}{\| {\mathcal H}\|}\cdot \frac{{\mathcal C_l}}{\| {\mathcal C_l}\|}\propto {\mathcal H}\cdot {\mathcal N_l}
\end{equation}
where $ {\mathcal H} \cdot {\mathcal C_l}$ is the dot product between ${\mathcal H}$ and a class hypervector ${\mathcal C_l}$, and ${\mathcal N_l}$ represents the normalized class hypervector, i.e., $\frac{\mathcal  C_l}{\| {\mathcal C_l}\|}$. Here $\| {\mathcal H}\|$ is a constant factor when comparing a query with all classes and thus can be eliminated. %, and $\|\overrightarrow {C_l}\|$ is a constant factor for a class so that only needs to be calculate once. 
The calculation of cosine similarity can hence be simplified to a dot product operation. 
For bipolar hypervectors, it is simplified to the Hamming distance. %$\overrightarrow{\mathcal{H}_1}\cdot \overrightarrow{\mathcal{H}_2}$
\textbf{Bundling (+):} element-wise addition of multiple hypervectors, e.g., ${\mathcal{H}}_{bundle}= {\mathcal{H}}_1 +{\mathcal{H}}_2$, generating a hypervector with the same dimension as inputs. In high-dimensional space, bundling works as a memory operation and provides an easy way to check the existence of a query hypervector in a bundled set. In the previous example,  $\delta({\mathcal{H}}_{bundle},{\mathcal{H}}_1)\gg 0 $ while $\delta({\mathcal{H}}_{bundle},{\mathcal{H}}_3)\approx 0 $, (${\mathcal{H}}_3\neq {\mathcal{H}}_1, {\mathcal{H}}_2)$.
\textbf{Binding (*):} element-wise multiplication associating two hypervectors to create another near-orthogonal hypervector, %new object in high-dimensional space that is orthogonal to all inputs, 
 i.e. ${\mathcal{H}}_{bind} ={\mathcal{H}}_{1}* {\mathcal{H}}_{2}.$ 
 %where $\delta(\overrightarrow{\mathcal{H}}_{bind}, \overrightarrow{\mathcal{H}}_{1})\approx 0$ and $\delta(\overrightarrow{\mathcal{H}}_{bind}, \overrightarrow{\mathcal{H}}_{2})\approx 0$. 
 Due to reversibility, in bipolar cases,
 %Due to the reversibility of this operation, 
 ${\mathcal{H}}_{bind} * {\mathcal{H}}_1 = {\mathcal{H}}_2$, information from both hypervectors can be preserved. 
  After generating all the encoded hypervectors %$\overrightarrow{\mathcal{H}}^l$ 
 of inputs for each class, %$l$, 
HDC training can take place. We elaborate our training framework in section \ref{sec:top2} and \ref{sec:regeneration}.
The inference phase of HDC consists of two steps: (\romannumeral 1) encode $(\encircle{A})$ inference data with the same encoder utilized in training to generate a query hypervector $ {\mathcal{Q}}$ $(\encircle{D})$, and (\romannumeral 2) calculate the distance or cosine similarity between $ {\mathcal{Q}}$ and each class hypervector $(\encircle{E})$. We then classify the query $ {\mathcal{Q}}$ to the class where it achieves the highest cosine similarity. $(\encircle{F})$.

\subsection{Top2-Classification}\label{sec:top2}
As explained in Section \ref{sec:intro}, SOTA HDC algorithms provide outstanding learning quality for top-2-classification while considerably weaker performance for top-1-classification. Inspired by this observation, our proposed $\Design$ first trains the model with an efficient and lightweight \textit{adaptive learning} algorithm (\encircle{G}, \encircle{H}), and utilize the partially trained model in each iteration to identify the top two most similar classes for every data point(\encircle{I}). In this way, we can identify (\encircle{N}) dimensions that mislead our model to select the incorrect labels and regenerate (\encircle{P}) those dimensions to enhance accuracy. 

\begin{algorithm}[t]
\small
  \caption{Adaptive Learning}\label{alg:adapt}
  \begin{algorithmic}[1]{}
    \Require{Training data points $\mathcal H (\mathcal H_1, \mathcal H_2, \ldots,  \mathcal H_n)$ with $q$ features and $k$ classes, labels for each data point $\mathcal L (\mathcal L_1,  \mathcal L_2, \ldots, \mathcal L_n)$, base vectors $\mathcal B$ ($\mathcal B_1, \mathcal B_2, \ldots, \mathcal B_q$) each with dimension $\mathcal D$, class hypervectors $\mathcal C$ ($\mathcal C_1, \mathcal C_2, \ldots \mathcal C_k$), learning rate $\eta$. }
    
    \Ensure{Class hypervectors $\mathcal C$ after one training iteration}. 
    \State $\mathcal H' = \mathcal H \cdot \mathcal B$ \textrm{ (matrix multiplication)}, \textit{dim}$(\mathcal H')=n\times \mathcal D$ \label{alg:adapt: encode}
    \ForEach{$\mathcal H_i \in \mathcal H'$}
        \State $\mathcal C_i = \max\{\delta(\mathcal H_i, \mathcal C_1), \delta(\mathcal H_i, \mathcal C_2), \ldots, \delta(\mathcal H_i, \mathcal C_k)\}$ \label{alg:adapt: cos}
        \If{$\mathcal L_i = \mathcal C_i$}
            \State continue
        \ElsIf{$\mathcal L_i \neq \mathcal C_i \land \mathcal L_i = \mathcal C_j (i\neq j)$}
            \State $\mathcal C_i \leftarrow \mathcal C_i -\eta\cdot[1-\delta(\mathcal H_i, \mathcal C_i ) ]\times \mathcal H_i$ \label{alg:adapt: update_begin}
            \State $\mathcal C_j \leftarrow \mathcal C_j +\eta\cdot[1-\delta(\mathcal H_i, \mathcal C_j ) ]\times \mathcal H_i$ \label{alg:adapt: update_end}
        \EndIf
    \EndFor
    \State \Return $\mathcal C$
  \end{algorithmic}
\end{algorithm}

\textbf{Adaptive Learning:}
As demonstrated in Algorithm \ref{alg:adapt}, our adaptive learning starts with encoding training data onto hyperdimensional space with a matrix multiplication of training data and base vectors (line \ref{alg:adapt: encode}). To reduce model saturation, we bundle encoded data points by scaling a proper weight to each of them depending on how much new information is added to class hypervectors. For instance, for a new encoded training sample $\mathcal H$, we update the model base on its cosine similarities (equation (\ref{eqn:cosine_similarity})) with all class hypervectors (line \ref{alg:adapt: cos}). % i.e., 
% \begin{equation}
% \small
%     \label{eqn:cosine_similarity}
%     \delta( {\mathcal H}, {\mathcal C_l})=\frac{ {\mathcal H} \cdot {\mathcal C_l}}{\| {\mathcal H}\|\cdot \| {\mathcal C_l}\|}=\frac{ {\mathcal H}}{\| {\mathcal H}\|}\cdot \frac{{\mathcal C_l}}{\| {\mathcal C_l}\|}\propto {\mathcal H}\cdot {\mathcal N_l}
% \end{equation}
% where $ {\mathcal H} \cdot {\mathcal C_l}$ is the dot product between ${\mathcal H}$ and a class hypervector ${\mathcal C_l}$ $(\encircle{G})$, and ${\mathcal N_l}$ represents the normalized class hypervector, i.e., $\frac{\mathcal  C_l}{\| {\mathcal C_l}\|}$. Here $\| {\mathcal H}\|$ is a constant factor when comparing a query with all classes and thus can be eliminated. %, and $\|\overrightarrow {C_l}\|$ is a constant factor for a class so that only needs to be calculate once. 
%The calculation of cosine similarity can hence be simplified to a dot product operation. 
If ${\mathcal H}$ has the highest cosine similarity with class $\mathcal L_i$ while it actually has label $\mathcal L_j$, the model updates $(\encircle{H})$ as %$\mathcal  C_j \leftarrow \mathcal  C_j + \eta(1-\delta_j)\times{\mathcal H}$ \textrm{ and } $\mathcal  C_{i} \leftarrow  \mathcal  C_i - \eta(1-\delta_{i})\times{\mathcal H}$, where $\eta$ is a learning rate (line \ref{alg:adapt: update_begin} - \ref{alg:adapt: update_end})
Algorithm \ref{alg:adapt} line \ref{alg:adapt: update_begin} - \ref{alg:adapt: update_end}. A large $\delta_l$, indicating the input data point is common or already exists in the model, updates the model by adding a very small portion of the encoded query ($1-\delta_l \approx 0$). In contrast, a small $\delta_l$, indicating a noticeably new pattern that is uncommon or does not already exist in the model, updates the model with a large factor ($1-\delta_l \approx 1$). 

\textbf{Top-2 Labels:} In every training iteration, after applying the adaptive learning algorithm, we utilize the partially trained model to identify the top two most similar labels (\encircle{I}) for each data point in order to identify the misleading dimensions in the next step (section \ref{sec:regeneration}). For instance, as demonstrated in Algorithm \ref{alg:undesired_dims}, for a given data point ${\mathcal H}$, we calculate its cosine similarity (equation (\ref{eqn:cosine_similarity})) with all class hypervectors (line \ref{alg:undesired_dims:cosine}). Suppose we get $\delta(\mathcal H, \mathcal C_i)>\delta(\mathcal H,\mathcal  C_j)>\ldots>\delta(\mathcal H, \mathcal C_k)$, then the top-2 labels for $\overrightarrow {\mathcal H}$ here are $\mathcal C_i$ and $\mathcal C_j$ (line \ref{alg:undesired_dims:class}).

%\vspace{-5mm}
\begin{algorithm}[t]
\small
  \caption{Identifying Undesired Dimensions}\label{alg:undesired_dims}
  \begin{algorithmic}[1]{}
    \Require{Encoded training data points $\mathcal H (\mathcal H_1, \mathcal H_2, \ldots,  \mathcal H_n)$ with dimension $\mathcal D$, correct labels $\mathcal L (\mathcal L_1,  \mathcal L_2, \ldots, \mathcal L_n)$ for all data points, class hypervectors $\mathcal C$ {($\mathcal C_1, \mathcal C_2, \ldots, \mathcal C_k$)}, weight parameters $\alpha, \beta, \theta (\theta<\beta)$}, regeneration rate $\mathcal R$. 
    
    \Ensure{Undesired dimensions $\mathcal U$ to drop}. 
    \ForEach{$\mathcal H_i \in \mathcal H$}
        \State $\Delta = \{\delta(\mathcal H_i, \mathcal C_1), \delta(\mathcal H_i, \mathcal C_2), \ldots, \delta(\mathcal H_i, \mathcal C_k)\}$\label{alg:undesired_dims:cosine}
        \State $\mathcal C_i = \max({\Delta}), \mathcal C_j = \max({\Delta} \backslash {\mathcal C_i})$\label{alg:undesired_dims:class}
        \If{$\mathcal L_i = \mathcal C_i$}\label{alg:undesired_dims:correct}
            \State continue
        \ElsIf{$\mathcal L_i = \mathcal C_j$}\label{alg:undesired_dims:partial_begin}
            \State $m = \lvert \mathcal H_i-\mathcal C_j\lvert, m_1 = \lvert \mathcal H_i-\mathcal C_i\lvert$ \textrm{ (elementwise)}\label{alg:undesired_dims:partial_distance}
            \State $\mathcal M_i = \alpha \cdot m-\beta \cdot m_1 $ \label{alg:undesired_dims:partial_end}
        \ElsIf{$(\mathcal L_i \neq \mathcal C_i) \land (\mathcal L_i \neq \mathcal C_j)$}
            \State {$n = \lvert \mathcal H_i - \mathcal L_i \lvert, n_1 = \lvert \mathcal H_i-\mathcal C_i\lvert, n_2 = \lvert \mathcal H_i-\mathcal C_j\lvert$ }\label{alg:undesired_dims:incorrect_distance}
            \State $\mathcal N_i =\alpha\cdot n_1+\beta \cdot n_2-\theta \cdot n $\label{alg:undesired_dims:incorrect_end}
        \EndIf
    \EndFor
    \State $\mathcal M = \{\mathcal M_1, \mathcal M_2, \ldots, \mathcal M_p\}, \mathcal N = \{\mathcal N_1, \mathcal N_2, \ldots, \mathcal N_{q}\}$ 
    \State $\mathcal M = \textrm{Normalize}(\mathcal M), \mathcal N = \textrm{Normalize}(\mathcal N)$
    \State $\mathcal M' = \textrm{sum}(\mathcal M, \textrm{columnwise}), \mathcal N' = \textrm{sum}(\mathcal N, \textrm{columnwise}) $
    \State $\mathcal U = \{\textrm{argsort}(\mathcal M')[0: \mathcal R\%\cdot \mathcal D]\}\cap \{\textrm{argsort}(\mathcal N')[0: \mathcal R\%\cdot\mathcal  D]\}$ \label{alg:undesired_dims:select}
    \State \Return $\mathcal U$
  \end{algorithmic}
\end{algorithm}

\subsection{Dimension Regeneration} \label{sec:regeneration}
\textbf{Identifying Undesired Dimensions:} 
In each training iteration, we separate results provided by \textit{top-2-classification} (section \ref{sec:top2}) into three categories: \textit{correct, partially correct}, and \textit{incorrect}. For instance, for a given data point $ {\mathcal H}$, suppose it has the highest cosine similarity with $\mathcal C_i$ and the second-highest cosine similarity score with $\mathcal C_j$. We classify the result as \textit{correct} if its true label is $\mathcal C_i$ and as \textit{partially correct} if its true label is $\mathcal C_j$. We classify the result as \textit{incorrect} if its true label is neither $\mathcal C_i$ nor $\mathcal C_j$. We then ignore data points classified as \textit{correct} in this iteration, and select the undesired dimensions utilizing those classified as \textit{partially correct} and \textit{incorrect} (\encircle{J}). As demonstrated in Algorithm \ref{alg:undesired_dims}, for each data point $\mathcal H$ classified as \textit{partially correct}, we calculate the distance of each dimension between its hypervector and its true label $\mathcal C_j$ with $\lvert \mathcal H - \mathcal C_j \lvert$ and $\mathcal C_i$, where it has the highest similarity score, with $\lvert \mathcal H - \mathcal C_i \lvert$, respectively (line \ref{alg:undesired_dims:partial_distance}). To identify dimensions of $\mathcal H$ that are closest to the wrong label $\mathcal C_i$ and farthest away from the true label $\mathcal C_j$,  we search for dimensions that maximize $\lvert \mathcal H - \mathcal C_j \lvert$ and minimize $\lvert \mathcal H - \mathcal C_i \lvert$. This is equivalent to search for dimensions that maximize both $\lvert \mathcal H - \mathcal C_j \lvert$ and $-\lvert \mathcal H - \mathcal C_i \lvert$. We thus define distance matrix $\mathcal M$, where each row vector is defined as $\mathcal M_i = \alpha \cdot \lvert  {\mathcal H} - \mathcal C_j\lvert -\beta \cdot \lvert  {\mathcal H} - \mathcal C_i\lvert $ where $\alpha$ and $\beta$ are weight parameters (line \ref{alg:undesired_dims:partial_end}). In this way, we effectively avoid selecting dimensions storing common information across the two classes, as eliminating these dimensions can potentially decrease classification accuracy for other data points. Similarly, for each data point $\mathcal H'$ marked as \textit{incorrect}, we calculate the distance of each dimension between its hypervector and its true label $\mathcal C_l$, $\mathcal C_i$, and $\mathcal C_j$, respectively (line \ref{alg:undesired_dims:incorrect_distance}). We then define a distance matrix $\mathcal N$  where each row is defined as $\mathcal N_i =\alpha\cdot  \lvert \mathcal H - \mathcal C_l \lvert -\beta \cdot \lvert \mathcal H' -\mathcal C_i\lvert -\theta \cdot \lvert \mathcal H'- \mathcal C_j\lvert$ with weight parameters $\alpha$, $\beta$ (line \ref{alg:undesired_dims:incorrect_end}), and $\theta$, aiming to search for dimensions that farthest from the true label and closest to the wrong labels $\mathcal C_i$ and $\mathcal C_j$(line \ref{alg:undesired_dims:partial_end}). After calculating both distance matrices $\mathcal M$ and $\mathcal N$, we normalize them and sum up each row vector for each matrix in a column-wise way to obtain two $1\times \mathcal D$ distance vectors $\mathcal M'$ and $\mathcal N'$ (line\ref{alg:undesired_dims:select}). To avoid over-eliminating dimensions, we only drop dimensions that have large values in both $\mathcal M'$ and $\mathcal N'$ (Fig. \ref{fig: flow}, \encircle{M}). We conduct this step by choosing the intersection part of the top $\mathcal R\%$ dimensions of $\mathcal M$ and $\mathcal N$ with the largest values (line \ref{alg:undesired_dims:select}), where $\mathcal R$ is the regeneration rate. 

\textbf{Dimension Regeneration:} 
To improve classification accuracy, $\Design$ regenerates those dimensions selected to drop $(\encircle{N})$, so that the new dimensions can potentially have a more positive impact on the classification and better differentiate patterns. For classification tasks, considering the non-linear relationship between features, we utilize an encoding method 
inspired by the Radial Basis Function (RBF)\cite{rahimi2007random}. Mathematically, for a feature vector $\mathcal F = \{f_1, f_2, \ldots, f_n\}(f_i\in \mathds{R}
)$ with $n$ features, we generate the corresponding hypervector $\mathcal H=\{h_1, h_2, \ldots, h_{\mathcal{D}}\} (0 \leq h_i \leq 1, h_i \in  \mathds{R})$ with ${\mathcal D}$ dimensions by calculating a dot product of feature vector with a randomly generated vector as $h_i = \cos(\mathcal B_i\cdot \mathcal F+c)\times \sin (\mathcal B_i\cdot \mathcal F)$, where $\mathcal B_i = \{b_1, b_2, \ldots, b_n\}$ is a randomly generated base vector with $b_i\sim \textit{Gaussian}(\mu =0,\sigma =1) \textrm{ and } c\sim \textit{Uniform}[0, 2\pi].$ % is a random value sampled uniformly from $[0, 2\pi]$. 
During regeneration, $\Design$ replaces the base vector of the selected dimension in the encoding module with another randomly generated vector from the Gaussian distribution. 

\textbf{Weight Parameters:} We define weight parameters $\alpha$, $\beta$, and $\theta$ when calculating the distance matrices $\mathcal M$ and $\mathcal N$ in Algorithm \ref{alg:undesired_dims}. $\alpha$ scales weights of dimensions being far away from correct labels while $\beta$ and $\theta$ provide weights for dimensions being close to wrong labels. 
Specifically, larger $\alpha$ values provide results with more \textit{sensitivity} by reducing the probability of a data sample being not classified to its true label, i.e. the false negative rate (FNR). In contrast, larger $\beta$ and $\theta$ values provide results with more \textit{specificity} by reducing the probability of a data sample being classified into wrong classes, i.e. the false positive rate (FPR). Mathematically, \textit{sensitivity} and \textit{specificity} are defined as:
\begin{equation*}
%\small
\begin{split}
    \textit{sensitivity} = \frac{\textit{True Positive}}{\textit{True Positive + False Negative}}=1-\textit{FNR}\\
    \textit{specificity} = \frac{\textit{True Negative}}{\textit{True Negative + False Positive}}=1-\textit{FPR}
    \end{split}
    %\vspace{-1mm}
\end{equation*}
The weight parameters can be adjusted according to the diverse needs of different learning tasks. We demonstrate trade-offs between \textit{sensitivity} and \textit{specificity} in section \ref{sec: accuracy}.

% \subsection{Inference}
% As demonstrated in Fig. \ref{fig: flow}, the inference phase of HDC consists of two steps: (\romannumeral 1) Encoding $(\encircle{A})$ inference data with the same encoder used in the training phase to generate a query hypervector $ {\mathcal{Q}}$ $(\encircle{D})$, and (\romannumeral 2) Computing the cosine similarity between $ {\mathcal{H}}$ and each class hypervector $(\encircle{E})$ in the same way as equation (\ref{eqn:cosine_similarity}). We then classify the query $ {\mathcal{Q}}$ to the class where it achieves the highest cosine similarity. $(\encircle{F})$. 

\section{Experimental Result} \label{sec:result}

\begin{table}[!t]
\fontsize{6.5}{3} \selectfont 
\caption{Datasets ($n$: number of features, $k$: number of classes)}\vspace{-0.3cm}
%\resizebox{1\columnwidth}{!}{
\vspace{-2mm}
\begin{center}
\begin{tabular}{l|crrrc} \toprule
  & $n$  & $k$  & 
  \begin{tabular}{@{}c@{}}\textbf{Train} \\ \textbf{Size}\end{tabular} 
  & \begin{tabular}{@{}c@{}}\textbf{Test} \\ \textbf{Size}\end{tabular} 
  & \textbf{Description} \\
\midrule
MNIST & 784  & 10	 & 60,000	 &	10,000 & Handwritten Recognition~\cite{ciregan2012multi}\\
UCIHAR & 561 & 12 &   6,213 &   1,554 & Mobile Activity Recognition~\cite{anguita2012human} \\
ISOLET & 617 & 26 & 6,238 & 1,559 & Voice Recognition\cite{lei2005half}\\
PAMAP2 & 54 &  5 &  233,687 & 115,101 &  Activity Recognition(IMU)~\cite{reiss2012introducing} \\
DIABETES & 49 & 3 & 66,000 & 34,000 & Outcomes of Diabetic Patients ~\cite{strack2014impact}\\
\bottomrule
\end{tabular}
\label{tb:datasets}
\end{center}%}
\vspace{-7mm}
\end{table}

\begin{figure*}[t]
%\vspace{-5mm}
  \centering
  {\includegraphics[width=\textwidth]{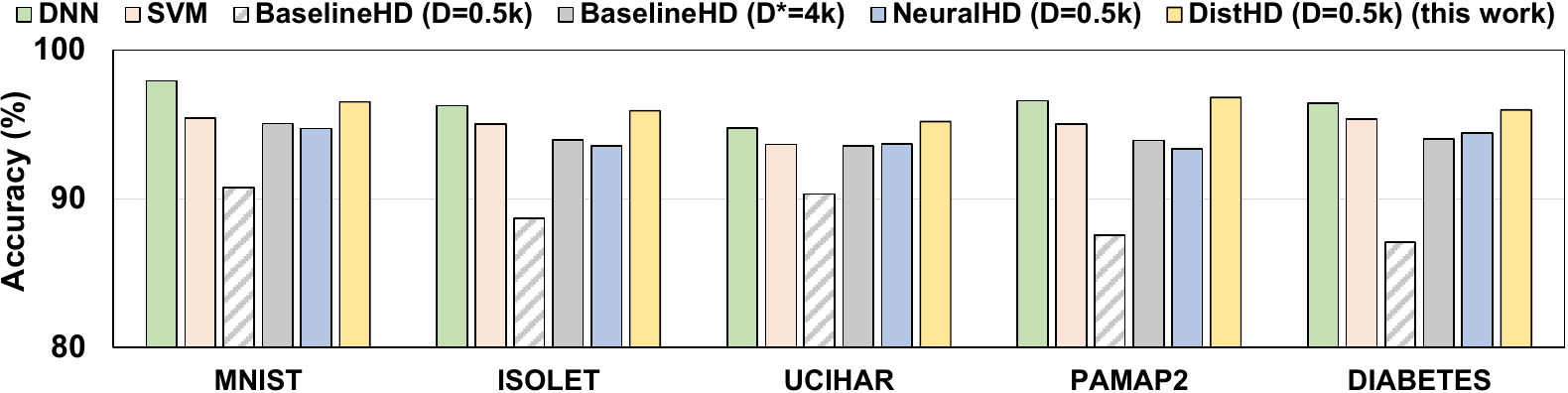}}
 \vspace{-6mm}
  \caption{Comparing Classification Accuracy of $\Design$ with State-of-the-art Learning Algorithms}
 \vspace{-1mm}
  \label{fig: accuracy} 
\end{figure*}

\begin{figure*}[t]
  \centering
  {\includegraphics[width=\textwidth]{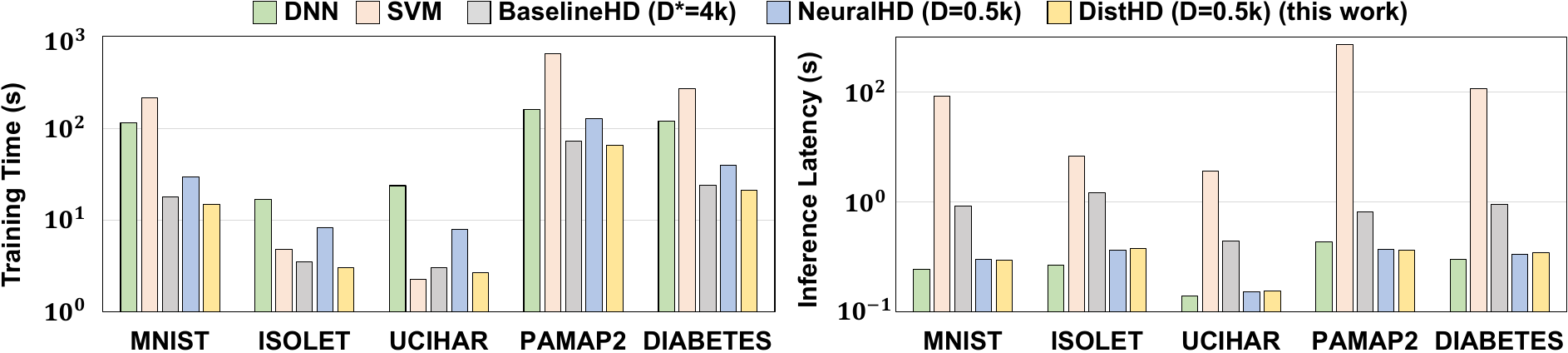}}
 \vspace{-6mm}
  \caption{Comparing Training and Inference Efficiency of $\Design$ with State-of-the-art Learning Algorithms}
 % \vspace{-2mm}
  \label{fig: efficiency} 
\end{figure*}

\begin{figure}[!t]
\centering
\vspace{-5mm}
\includegraphics[width=\linewidth]{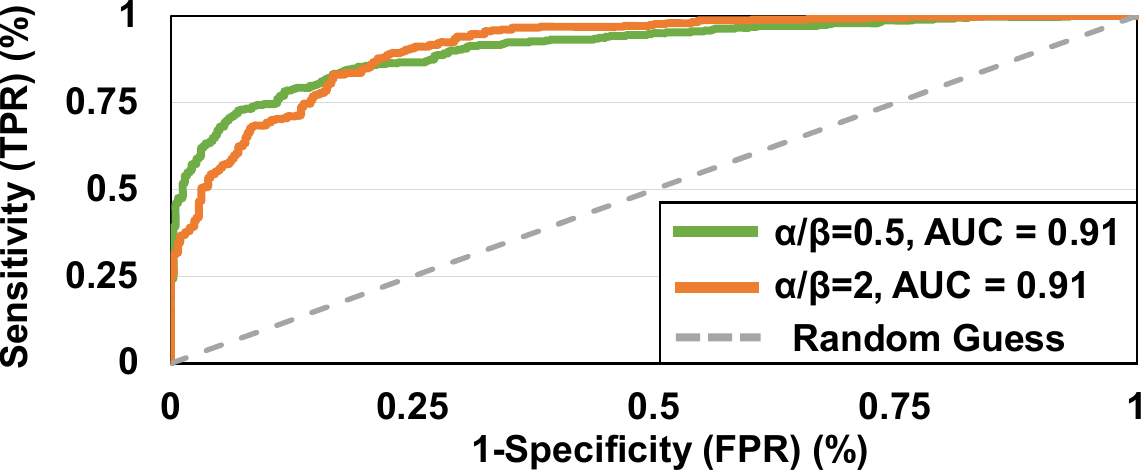}
\vspace{-6mm}
\caption{ROC curve of $\Design$ using different weight parameters $\alpha$ and $\beta$}
% \vspace{-5mm}
\label{fig: roc}  
\end{figure}

\subsection{Experimental Setup} 
We evaluated the effectiveness of our proposed $\Design$ learning framework with CPU (Intel Core i9-12900) on widely-used machine learning datasets listed in TABLE \ref{tb:datasets}. The $\Design$ code was written in Python with NumPy and optimized for performance. We compare $\Design$ with state-of-the-art (SOTA) DNNs~\cite{taud2018multilayer}, SVMs~\cite{hearst1998support}, SOTA HDC algorithms~\cite{rahimi2007random} in terms of accuracy, training and inference efficiency, and robustness against hardware noise. We also compare $\Design$ with NeuralHD~\cite{zou2021scalable}, a recently proposed dynamic encoding technique for HDC aiming at dimension reduction. 

\subsection{DistHD Accuracy}\label{sec: accuracy}

\textbf{DistHD vs. SOTA ML Algorithms:} We compare the classification accuracy of $\Design$ with SOTA learning algorithms, including 
SOTA deep neural networks (DNNs) and support vector machines (SVMs). The SOTA DNN algorithm is trained with TensorFlow~\cite{taud2018multilayer} while SVM is trained with the scikit-learn library~\cite{hearst1998support}. We utilize the common practice of grid search to identify the best hyper-parameters for each model. As demonstrated in Fig. \ref{fig: accuracy}, $\Design$ provides comparable accuracy to SOTA DNNs and 1.17\% higher accuracy than SVMs.

\begin{figure}[!t]
\centering
\vspace{-2mm}
\includegraphics[width=\linewidth]{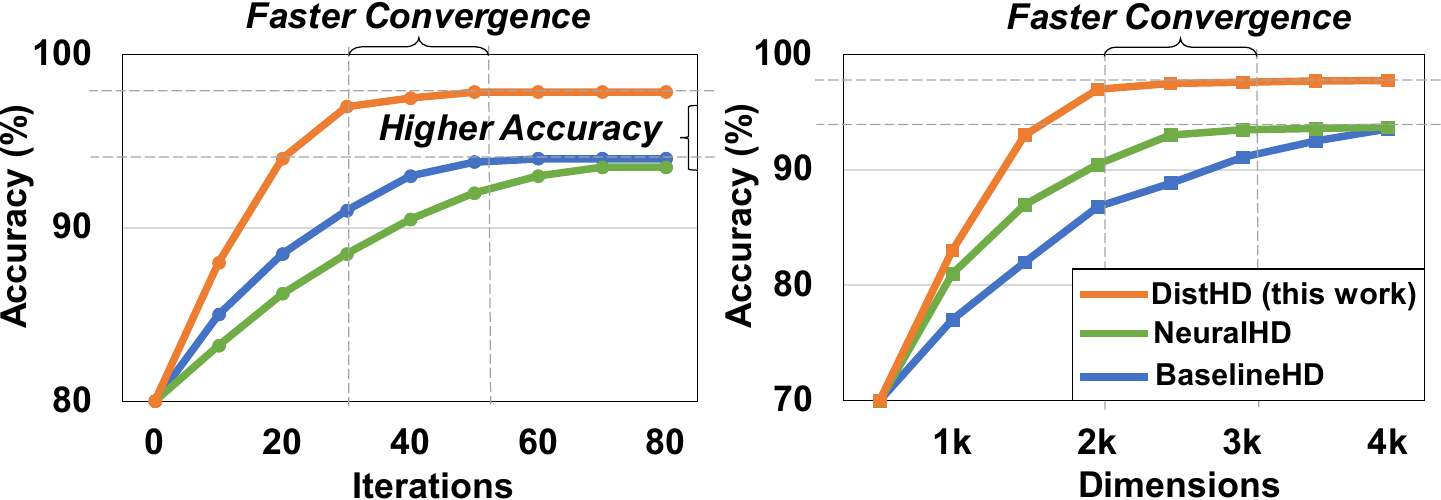}
\vspace{-6mm}
\caption{Comparing Convergence Speed of $\Design$ with other HDC algorithms}
\vspace{-5mm}
\label{fig: converege}  
\end{figure}

\textbf{DistHD vs. SOTA HDCs:}
We compare the accuracy of $\Design$ with SOTA HDC algorithms that are incapable of regenerating dimensions (baselineHD)~\cite{rahimi2016robust} and a recently proposed HDC learning approach using a dynamic encoder (NeuralHD)~\cite{zou2021scalable}. The results of baselineHD are reported in two dimensionality: (\romannumeral 1) \textit{Physical dimensionality} ($\mathcal D=0.5$\textrm{k}) of NeuralHD and $\Design$, a compressed dimensionality designed for resource-constrained computing devices. (\romannumeral 2) \textit{Effective dimensionality} ($\mathcal D^*=4\textrm{k}$), defined as the sum of the physical dimensions $(\mathcal D)$ of $\Design$ with the regenerated dimensions throughout the retraining iterations. Mathematically, $\mathcal D^*=\mathcal D+\mathcal D \times \mathcal R\% \times \textit{Number of Iterations}$, where $\mathcal R$ is the regeneration rate. We train each HDC model until it reaches convergence. As shown in Fig. \ref{fig: converege}, baselineHD and NeuralHD converge at lower accuracy than DistHD due to lacking the capability to fully utilize the information learned during the training process. As demonstrated in Fig. \ref{fig: accuracy}, $\Design$ ($\mathcal D=0.5$\textrm{k}) delivers on average $6.96\%$ and $1.88\%$ higher accuracy than baselineHD ($\mathcal D=0.5\textrm{k}$) and NeuralHD ($\mathcal D=0.5\textrm{k}$), respectively. Additionally, $\Design$ achieves $1.82\%$ higher accuracy than baselineHD ($\mathcal D^*=4\textrm{k}$). This indicates that $\Design$ is capable of outperforming SOTA HDC in terms of accuracy while reducing physical dimensionality by $8.0\times$ on average. 

\textbf{Sensitivity vs. Specificity: } We present trade-offs between sensitivity and specificity using ROC curves and area under ROC curves (AUC) in Fig. \ref{fig: roc}. For two groups of parameters showing comparable accuracy and AUC, with the decrease of the specificity, the model with larger $\alpha$ shows a sharper increase in sensitivity and is more likely to provide higher sensitivity for classification tasks. In contrast, the model with larger $\beta$ loses less specificity with the increase of sensitivity and is more likely to deliver results with higher specificity. We can tune our weight parameters according to the diverse needs of learning tasks for the best outcomes. 
% the same physical dimensionality $(D=0.5\textrm{k})$ as $\Design$, and the same effective dimensionality $(D^* = 4\textrm{k})$ as $\Design$. The effective dimensionality $(D^*)$ is defined as the addition of the physical dimensions $(D)$ of $\Design$ with the regenerated dimensions throughout the retraining iterations. Mathematically, $D^*=D+D \times R \times \textit{Number of Iterations}$, where $R$ is the regeneration rate. 
\vspace{-2mm}
\subsection{DistHD Efficiency}
For fairness, we compare the training and inference efficiency of the SOTA DNN, SVM, baselineHD ($\mathcal D^* = 4\textrm{k}$), NeuralHD ($\mathcal D = 0.5\textrm{k}$), and $\Design$ ($\mathcal D = 0.5\textrm{k}$) as they achieve comparable accuracy according to Fig. \ref{fig: accuracy}. As shown in Fig. \ref{fig: efficiency}, SVMs take a significantly longer time for both training and inference for large datasets such as PAMAP and DIABETES. $\Design$ delivers considerably higher efficiency than SOTA DNNs ($5.97\times$ faster training,  comparable inference latency), SVMs, and SOTA HDC ($1.15\times$ faster training,  $8.09\times$ faster inference). $\Design$ also delivers a $2.32\times$ speedup in training compared to NeuralHD. $\Design$ achieves such high training efficiency due to its capability to reach convergence with noticeably fewer iterations and lower dimensionality than other HDC algorithms, as demonstrated in Fig. \ref{fig: converege}. Additionally, $\Design$ delivers short inference latency since it requires significantly lower dimensionality, effectively accelerating the process of encoding query vectors and calculating similarity scores.

\subsection{Robustness of DistHD against Hardware Noises}
\begin{figure}[!t]
\centering
%\vspace{-3mm}
\includegraphics[width=\linewidth]{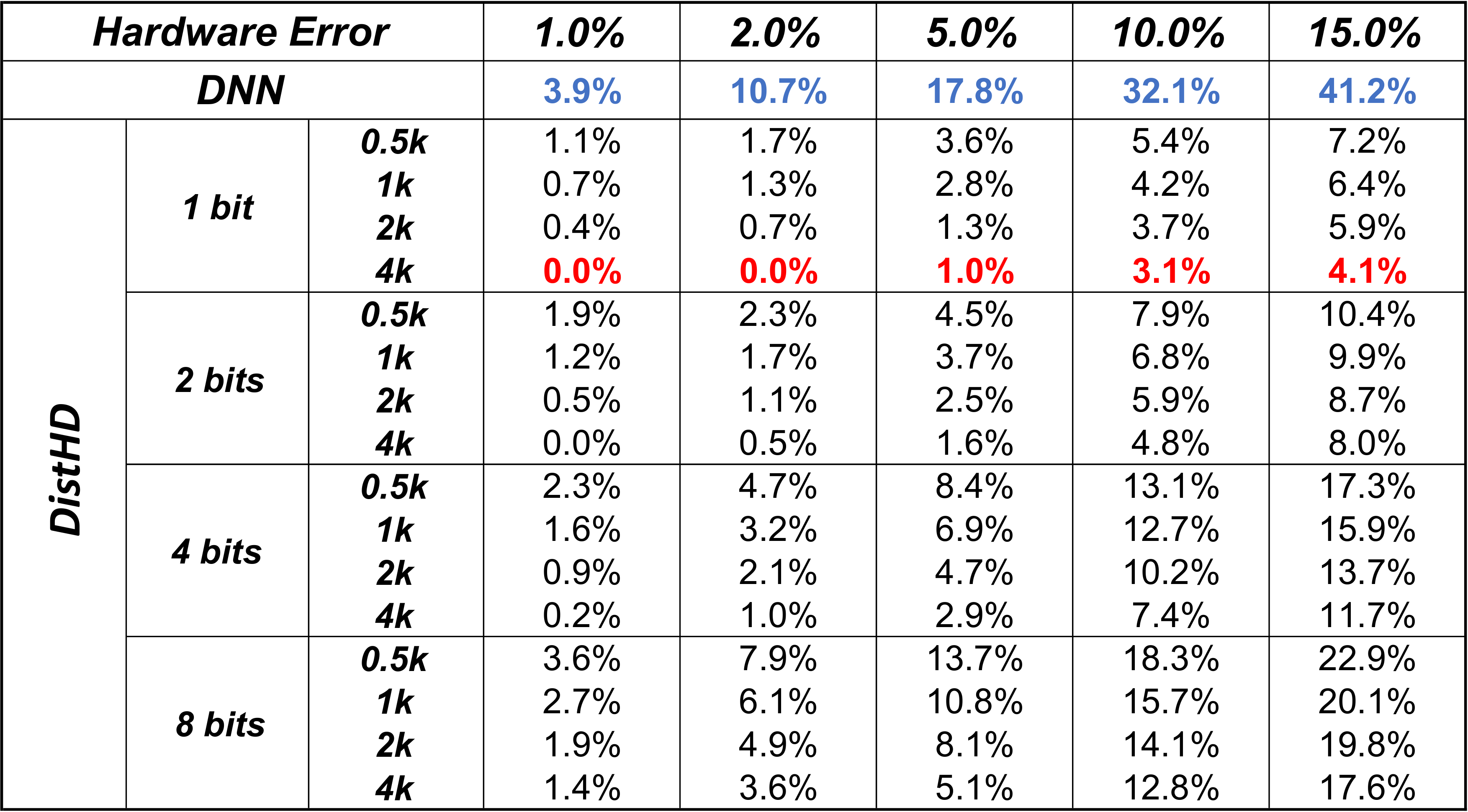}
\vspace{-6mm}
\caption{Comparing Quality Loss of $\Design$ with DNNs}
\vspace{-5mm}
\label{fig: robust}  
\end{figure}

% Table \ref{} the robustness of DNN and $\Design$ against hardware noises. For $\Design$, the results are reported using vectors with different precision. The dimensionality of the vector is selected when $\Design$ provides maximum accuracy. Our evaluation shows that $\Design$ provides significantly higher roubustness to noise as compared to DNN. In DNNs, the weights are represented in 8-bit values 
One of the main advantages of $\Design$ is its high robustness against noise and failure. In $\Design$,  %hypervectors are random and holographic with i.i.d. components. E
each hypervector stores information across all its components so that no component is more responsible for storing any more information than another, making each hypervector robust against errors. Here we compare the robustness of $\Design$ and DNN to hardware noise in Fig. \ref{fig: robust} by showing the average quality loss of the DNN and $\Design$ under different percentages of hardware errors. The error rate refers to the percentage of random bit flips on memory storing DNN and $\Design$ models. For fairness, all DNN weights are quantized to their effective 8-bit representation. In DNN, random bit flip results in significant quality loss as corruptions on most significant bits can cause major weight changes. In contrast, $\Design$ provides significantly higher robustness to noise due to its redundant and holographic distribution of patterns in high-dimensional space. Additionally, all dimensions equally contribute to storing information, and thus failure on partial data will not result in the loss of entire information. 

$\Design$ demonstrates the maximum robustness using hypervectors with $4\textrm{k}$ dimensions in 1-bit precision, that is on average $12.90\times$ higher robustness than DNN. Increasing precision lowers the robustness of $\Design$ since random flips on more significant bits will introduce more loss of accuracy. For instance, for $10\%$ bit flips in hardware, $\Design$ using 1-bit precision and $4\textrm{k}$ dimensions provides $10.35\times$ and $4.13\times$ higher robustness than DNN and $\Design$ using 8 bits with the same dimensionality, respectively. Additionally, higher dimensionality improves the robustness of $\Design$ against noise due to more redundant and holographic information distribution. For example, for $10\%$ hardware error, $\Design$ using $4\textrm{k}$ dimensions and 8-bit precision achieves $1.43\times$ higher robustness than $\Design$ using $0.5\textrm{k}$ dimensions with the same bitwidth. %On average, CyberHD achieves a $4.65\times$ higher robustness compared to the state-of-the-art DNN algorithm. 

% \begin{table}[!t]
% %\small
% %\vspace{-6mm}
% %\footnotesize
% \caption{Quality Loss Using Noisy Hardware}\vspace{-0.3cm}
% \resizebox{1\columnwidth}{!}{
% \vspace{-2mm}
% \begin{center}
% \begin{tabular}{l|ccccc} \toprule
%  \textit{Hardware Error} & $n$  & $k$  & 
%   \textbf{1\%} & \textbf{2\%} & \textbf{5\%} & \textbf{10\%} & \textbf{15\%}
% \midrule
% MNIST & 784  & 10	 & 60,000	 &	10,000 & Handwritten Recognition~\cite{ciregan2012multi}\\
% UCIHAR & 561 & 12 &   6,213 &   1,554 & Mobile Activity Recognition~\cite{anguita2012human} \\
% ISOLET & 617 & 26 & 6,238 & 1,559 & Voice Recognition\cite{lei2005half}\\
% PAMAP2 & 54 &  5 &  233,687 & 115,101 &  Activity Recognition(IMU)~\cite{reiss2012introducing} \\
% DIABETES & 49 & 3 & 66,000 & 34,000 & Outcome of Diabetic Patients ~\cite{strack2014impact}\\
% \bottomrule
% \end{tabular}
% \label{tb:robustness}
% \end{center}}
% \vspace{-5mm}
% \end{table}

\section{Conclusion} \label{sec:conclusion}
In this paper, we propose $\Design$, an accurate, efficient, and robust HDC learning framework. With a powerful dynamic encoding technique, $\Design$ identifies and regenerates dimensions that mislead the classification and reduce the learning accuracy. Our evaluations on a wide range of machine learning datasets demonstrate that $\Design$ delivers on average $2.12\%$ higher accuracy than SOTA HDC algorithms and reduces the dimensionality by $8.0\times$. It also significantly outperforms SOTA DNNs and HDCs in terms of both training and inference efficiency. Additionally, the holographic distribution of patterns in high dimensional space provides $\Design$ with $12.90\times$ higher robustness than SOTA DNNs. The performance of $\Design$ makes it an outstanding solution for edge platforms.

\section{Acknowledgements}

This work was supported in part by National Science Foundation \#2127780, Semiconductor Research Corporation (SRC), Office of Naval Research, grants \#N00014-21-1-2225 and \#N00014-22-1-2067, the Air Force Office of Scientific Research under award \#FA9550-22-1-0253, and generous gifts from Xilinx and Cisco.

% state-of-the-art HDC algorithms and DNNs in both training and inference efficiency ($1.85\times$ faster training and $15.29\times$ faster inference than HDC, $2.47\times$  faster training and $2.20\times$ faster inference than DNN), while achieving comparable accuracy. 
%% Specifically, $\Design$ performs on average $1.85\times$ and $15.29\times$ faster than state-of-the-art HDC algorithms in training and inference, respectively. 

\bibliographystyle{unsrt}
\bibliography{reference}

\begin{thebibliography}{10}

\bibitem{chen2019deep}
Jiasi Chen and Xukan Ran.
\newblock Deep learning with edge computing: A review.
\newblock {\em Proceedings of the IEEE}, 2019.

\bibitem{shi2016edge}
Weisong Shi et~al.
\newblock Edge computing: Vision and challenges.
\newblock {\em IEEE Internet of Things journal}, 2016.

\bibitem{pan2017future}
Jianli Pan et~al.
\newblock Future edge cloud and edge computing for internet of things
  applications.
\newblock {\em Internet of Things Journal}, 2017.

\bibitem{ge2020classification}
Lulu Ge and Keshab~K Parhi.
\newblock Classification using hyperdimensional computing: A review.
\newblock {\em IEEE Circuits and Systems Magazine}, 2020.

\bibitem{imani2019framework}
Mohsen Imani, Yeseong Kim, et~al.
\newblock A framework for collaborative learning in secure high-dimensional
  space.
\newblock In {\em CLOUD}. IEEE, 2019.

\bibitem{rahimi2016robust}
Abbas Rahimi et~al.
\newblock A robust and energy-efficient classifier using brain-inspired
  hyperdimensional computing.
\newblock In {\em ISLPED}, 2016.

\bibitem{zou2021scalable}
Zhuowen Zou et~al.
\newblock Scalable edge-based hyperdimensional learning system with brain-like
  neural adaptation.
\newblock In {\em SC}, 2021.

\bibitem{andersen2003aging}
Birgitte~Bo Andersen et~al.
\newblock Aging of the human cerebellum: a stereological study.
\newblock {\em Journal of Comparative Neurology}, 2003.

\bibitem{warden2019tinyml}
Pete Warden et~al.
\newblock {\em TinyML}.
\newblock O'Reilly Media, Incorporated, 2019.

\bibitem{david2021tensorflow}
Robert David et~al.
\newblock Tensorflow lite micro: Embedded machine learning for tinyml systems.
\newblock {\em Proceedings of MLSys}, 2021.

\bibitem{sakr2020machine}
Fouad Sakr, Francesco Bellotti, Riccardo Berta, and Alessandro De~Gloria.
\newblock Machine learning on mainstream microcontrollers.
\newblock {\em Sensors}, 2020.

\bibitem{xcubeai}
{X-Cube-AI: AI expansion pack for STM32CubeMX}.
\newblock \url{https://www.st.com/en/embedded-software/x-cube-ai.html}.

\bibitem{ko2018edge}
Jong~Hwan Ko et~al.
\newblock Edge-host partitioning of deep neural networks with feature space
  encoding for resource-constrained internet-of-things platforms.
\newblock In {\em AVSS}. IEEE, 2018.

\bibitem{bonawitz2017practical}
Keith Bonawitz et~al.
\newblock Practical secure aggregation for privacy-preserving machine learning.
\newblock In {\em ACM SIGSAC CCS}, 2017.

\bibitem{li2020federated}
Tian Li et~al.
\newblock Federated learning: Challenges, methods, and future directions.
\newblock {\em Signal Processing Magazine}, 2020.

\bibitem{luo2022keepedge}
Haoyu Luo et~al.
\newblock Keepedge: A knowledge distillation empowered edge intelligence
  framework for visual assisted positioning in uav delivery.
\newblock {\em Transactions on Mobile Computing}, 2022.

\bibitem{poduval2022graphd}
P.~Poduval et~al.
\newblock {GrapHD}: Graph-based hyperdimensional memorization for brain-like
  cognitive learning.
\newblock {\em Frontiers in Neuroscience}, 2022.

\bibitem{burrello2019laelaps}
Alessio Burrello et~al.
\newblock Laelaps: An energy-efficient seizure detection algorithm from
  long-term human ieeg recordings without false alarms.
\newblock In {\em DATE}. IEEE, 2019.

\bibitem{kim2018efficient}
Yeseong Kim et~al.
\newblock Efficient human activity recognition using hyperdimensional
  computing.
\newblock In {\em IoT}, 2018.

\bibitem{imani2017exploring}
Mohsen Imani et~al.
\newblock Exploring hyperdimensional associative memory.
\newblock In {\em HPCA}. IEEE, 2017.

\bibitem{rahimi2007random}
Ali Rahimi and Benjamin Recht.
\newblock Random features for large-scale kernel machines.
\newblock {\em Advances in neural information processing systems}, 2007.

\bibitem{ciregan2012multi}
Dan Ciregan, Ueli Meier, and J{\"u}rgen Schmidhuber.
\newblock Multi-column deep neural networks for image classification.
\newblock In {\em CVPR}. IEEE, 2012.

\bibitem{anguita2012human}
Davide Anguita et~al.
\newblock Human activity recognition on smartphones using a multiclass
  hardware-friendly support vector machine.
\newblock In {\em IWAAL}. Springer, 2012.

\bibitem{lei2005half}
Hansheng Lei et~al.
\newblock Half-against-half multi-class support vector machines.
\newblock In {\em International Workshop on Multiple Classifier Systems}.
  Springer, 2005.

\bibitem{reiss2012introducing}
Attila Reiss et~al.
\newblock Introducing a new benchmarked dataset for activity monitoring.
\newblock In {\em ISWC}. IEEE, 2012.

\bibitem{strack2014impact}
Beata Strack et~al.
\newblock Impact of hba1c measurement on hospital readmission rates: analysis
  of 70,000 clinical database patient records.
\newblock {\em BioMed research international}, 2014.

\bibitem{taud2018multilayer}
Hind Taud et~al.
\newblock Multilayer perceptron (mlp).
\newblock In {\em Geomatic approaches for modeling land change scenarios}.
  Springer, 2018.

\bibitem{hearst1998support}
Marti~A. Hearst et~al.
\newblock Support vector machines.
\newblock {\em Intelligent Systems and their applications}, 1998.

\end{thebibliography}
\end{document}